\documentclass{article} 
\usepackage{tree-dvips} 
\usepackage{a4} 
\begin{document} 
 
\renewcommand{\baselinestretch}{1.2} 
\newcommand{\mysection}[1]{\refstepcounter{section} 
                           \section*{{\protect{\mbox{\large\bf  
                                        \thesection\ \ #1}}}}} 
 
\newcommand{\mysubsection}[1]{\refstepcounter{subsection} 
                             \subsection*{\protect{\mbox{\normalsize\bf  
                             \thesubsection\ \ #1}}}\vspace{-6pt}} 
\newcommand{\mysubsubsection}[1] 
                       {\subsubsection*{\protect{\mbox{\normalsize\bf 
                               #1}}}}

\vspace*{2.4cm} 
\begin{center} 
{\Large\bf GENERIC\ RULES\ AND\ NON-CONSTITUENT\ COORDINATION\ } 
\footnote{Appeared in {\em Proceedings of the Fourth International
    Workshop on Parsing Technologies}, Prague, September 1995,
  pp. 99-110. Also as cmp-lg/9601002 .}
\vspace{1cm} 
 
\begin{tabular}{cc} 
{\Large Julio Gonzalo}  & {\Large Teresa Sol\'{\i}as} \\  
{\large UNED} & {\large Universidad de Valladolid} \\ 
{\large \texttt{julio@ieec.uned.es}} & {\large
  \texttt{solias@cpd.uva.es}}  
\end{tabular} 
\end {center} 
 
\vspace*{1cm} 
 
\begin{center} 
\centerline{{\small {\bf Abstract}}}  
 
\begin{tabular}{ @{\hspace{.5cm}} p{13cm}  @{\hspace{.5cm}}   } 
{\small  
We present a metagrammatical formalism, {\em generic rules}, 
to give a default interpretation to grammar 
rules. Our formalism introduces a process of {\em dynamic binding} 
interfacing the level of pure grammatical knowledge representation and 
the parsing level. We present an approach to 
non-constituent coordination within categorial grammars, and 
reformulate it as a generic rule. This reformulation is context-free 
parsable and reduces drastically the search space associated to the 
parsing task for such phenomena. 
} 
\end{tabular} 
\end{center} 
 
\newcommand{\ccaption}[1]{\caption{{\small #1}}} 
 
\newcommand{\phrase}[1]{`{\em #1}'} 
\newcommand{\rem}[1]{{\footnotesize\bf #1}} 
 
\newcommand{\tuple}[1]{\mbox{$\langle #1 \rangle$}} 
 
\newcommand{\type}[1]{\mbox{\texttt{\textit{#1}}}} 
 
\newcommand{\sem}[1]{\mbox{#1}}

\newcommand{\csymbol} 
  {\otimes} 
 
\newcommand{\ctype}[2] 
  {\mbox{$\type{#1} \! \! \csymbol  \! \!\type{#2}$}} 
 
\newcommand{\cctype}[2] 
  {\mbox{$#1 \! \! \csymbol \! \! #2$ }} 
 
\newcommand{\vv}[1] 
  {\cctype{\type{#1}_{1}}{\type{#1}_{2}}}  
 
\newcommand{\set}[1] 
  {{\cal #1}} 
 
\newcommand{\cset}[1] 
  {{\cal #1}^{*}} 
 
\newcommand{\cproduct}[1] 
  {\set{#1}\times\set{#1}} 
 
\newcommand{\corder} 
  {\preceq} 
 
\newcommand{\poset}[1] 
   {\langle \leq , {\cal #1} \rangle} 
 
\newcommand{\cposet}[1] 
   {\langle \corder ,{\cal #1}^{*} \rangle} 
 
\newcommand{\grule}[1] 
{\mbox{#1}} 
 
\newcommand{\prule}[3] 
   {\mbox{#1}_{ \ctype{#2}{#3}}} 
 
\newcommand{\pprule}[3] 
   {\mbox{#1}_{ \cctype{#2}{#3}}} 
 
\newcommand{\prulef}[2] 
   {\pprule{#1}{\type{#2}_{1}}{\type{#2}_{2}}}

\newcommand{\Prule}[4] 
   {\prule{#1}{#2}{#3}(x,y) = \fun{#4} (x,y)} 
 
\newcommand{\PPrule}[4] 
   {\pprule{#1}{#2}{#3}(x,y) = #4 (x,y)} 
 
\newcommand{\fun}[1] 
   {\mbox{#1}} 
 
\newcommand{\cfg}[3] 
  {\type{#1} \ra \type{#2}\type{#3}}

 
\newcommand{\nil}{\perp} 
 
 
\newcommand{\dbfun}[1]  
{\mbox{$\mathcal{#1}$}} 
 
\newcommand{\precorder}[1] 
{\corder_{#1}} 
 
\newcommand{\precede}[3] 
{\mbox{$#1 \precorder{#3} #2$}} 
 
\newcommand{\myoneover}[2]{\begin{array}{@{}c@{}} #1 \\ 
 #2  \end{array}} 
 
\newcommand{\rasgo}[1] 
{\mbox{@1}} 
 
\newcommand{\EVRECUADRO}[2]{ 
\begin{tabular}{ | @{\hspace{.5cm}} p{#1}  @{\hspace{.5cm}} |  } 
\hline 
\\ 
#2 
\\ \\ 
\hline 
\end{tabular}} 
 
 
\newcommand{\comp}[2]{ #1 \circ #2} 
\newcommand{\disj}[2]{#1 \vee #2} 
\newcommand{\opt}[1]{[#1]} 
\newcommand{\specialconj}{\mbox{$\type{c}_{\tuple{}}$}} 
\newcommand{\fa}{\mbox{{\bf fa}}} 
\newcommand{\ba}{\mbox{{\bf ba}}} 
\newcommand{\scan}{\mbox{{\bf scan}}} 
\newcommand{\Ituple}{\mbox{$\mathbf{I_{\langle \cdots \rangle}}$}} 
\newcommand{\Dtuple}{\mbox{$\mathbf{D_{\langle \cdots \rangle}}$}} 
 

\newcommand{\oneover}[2]{\begin{array}{@{}c@{}} #1 \\
\hline #2  \end{array}}
\newcommand{\seq}[1]{\langle #1 \rangle}
\newcommand{\ra}{\rightarrow}
\newcommand{\Ra}{\Rightarrow}
\newcommand{\derivestar}{\stackrel{*}{\Rightarrow}}

\newcommand{\bk}{\mbox{$\backslash$}}


\mysection{Introduction}

Grammatical theories always make a distinction between unmarked and
marked phenomena. The general case is to use certain mechanisms which
are likely to deal with regular behaviour. More complex mechanisms
should be available when exceptional behaviour is present. However,
this difference is not formally expressed in most grammar accounts of
complex phenomena. The need to restrict licensing of additional
grammar resources is commonly asserted outside of the formal framework
used. 
As exceptional rules involve a higher computational cost, parsing
processes associated to such formalisms become computationally
intractable. 

Non-constituent coordination treatments show this pattern. In general, 
a conjunction combines constituents. But there are cases where the 
conjunction coordinates linguistic elements which do not form a 
constituent. For instance, consider  
\phrase{John went to Alabama in summer and to Spain in fall}. 
The conjunction \phrase{and} is coordinating \phrase{ to Alabama in 
summer} with \phrase{ to Spain in fall}, which do not form
constituents. 
The solutions displayed in the literature incorporate additional 
grammatical resources to deal with such data. 
We would expect a high restriction 
on the use of the rules associated to this phenomena,  
but all we have is informal comments about when to apply them. 
That makes parsing time of the grammar exponential 
in the length of the string.  
 
Though there is a general concern on the 
need to make explicit some process of rule licensing, no formal 
framework has  
been proposed, to our knowledge, to regulate the  interaction between 
regular and exceptional grammatical resources.  
In this paper we propose a metagrammatical formalism, {\em generic 
rules}, to give a different degree of specificity to each grammatical 
rule. We present an approach to parse non-constituent coordination 
within categorial grammars that can be parsed in polynomial time when 
reformulated as a generic rule.  
 
The essential idea is to introduce a process of {\em dynamic binding}
interfacing the level of pure grammatical representation and the
parsing processes.  When the parsing process calls the grammar to
combine linguistic objects, the dynamic binding process inspects the
applicable grammatical resources, considers their different priority
and returns an effective rule, or set of rules, to be applied on such
linguistic objects.  This approach is inspired in the object-oriented
programming paradigm, where polymorphism is exploited in order to get
default and specific behaviour.
 
In order to make dynamic binding a meaningful process, we will view
generic and specific behaviour as object-oriented functions that map
daughters information into mother features.  A set of such functions
will be called a {\em generic rule}. Precedence between them will be
established according to the specificity of the type constraints on
the daughters.
 
This paper is structured as follows: first, we outline the main ideas 
behind our approach. Next, we make a more formal presentation of {\em 
  generic rules}. Then, we turn to the linguistic problem of 
non-constituent coordination, and give a linguistic explanation within 
categorial grammar that uses the {\em tuple} operation introduced in 
\cite{Solias-93,Morrill-93}. Recognition with that grammar is NP-hard, 
as it is with most extensions of Lambek Calculus. We present, then, a 
reformulation of that grammar as a generic rule. The representational 
facilities of generic rules allow the specification of each 
grammatical resource at an appropriate level of specificity, 
drastically constraining the search space for the parsing task. 
 
\mysection{Generic Rules}
\label{generic-rules}

\mysubsection{Underlying Ideas}
\label{ideas}

Consider a context-free grammar - possibly augmented with functional
restrictions attached to each rule - that includes the hierarchy
$\texttt{VP} \ra \texttt{VP}_{1} \ | \ \texttt{VP}_{2} \ | \ 
\texttt{VP}_{3} \ | \ \texttt{VP}_{4} \ | \ \texttt{VP}_{i}$ and the
rule $\texttt{S} \ra \texttt{NP} \ \texttt{VP}$.

If we decide to write down a new rule that specifies a different, more
specific behaviour for $\texttt{VP}_{i}$ when combining with a
$\texttt{ NP}$, we could think of adding a new rule as:
$\texttt{S}_{i} \ra \texttt{NP} \ \texttt{VP}_{i}$ .  If our aim is to
state this rule as {\em the} rule to be used when the \texttt{VP}
belongs to the more specific subclass $\texttt{VP}_{i}$, we want the
production $\texttt{S} \Rightarrow \texttt{NP} \ \texttt{VP}_{i}$ to
be valid, but not the production $\texttt{S} \Rightarrow \texttt{NP} \ 
\texttt{VP} \Rightarrow \texttt{NP} \ \texttt{VP}_{i}$.  Just adding
$\texttt{S}_{i} \ra \texttt{NP} \ \texttt{VP}_{i}$ to the CFG does not
capture the exceptional sense of the rule; in a CFG, both derivations
are equally possible.

We have two options to get the desired behaviour from the CFG.  The
first one is to keep $\texttt{S} \ra \texttt{NP} \ \texttt{VP}$ and
rewrite the specific rule $\texttt{S}_{i} \ra \texttt{NP} \ 
\texttt{VP}_{i}$ into the following set of rules:

\begin{equation}
  \texttt{S} \ra \texttt{NP} \ \texttt{VP}_{1} \ | \ \texttt{NP} \ 
  \texttt{VP}_{2} \ | \ \texttt{NP} \ \texttt{VP}_{3} \ | \ 
  \texttt{NP} \ \texttt{VP}_{4}
\end{equation}

The second one is to rewrite the hierarchy, in order to keep the
number of rules invariant:

\begin{equation}
\begin{array}{ll}
  \texttt{VP} \ra \texttt{VP}_{i} \ | \ \texttt{VP}_{t} & \texttt{S}
  \ra \texttt{NP} \ \texttt{VP}_{t}\\ 
  \texttt{VP}_{t} \ra
  \texttt{VP}_{1} \ | \ \texttt{VP}_{2} \ | \ \texttt{VP}_{3} \ | \ 
  \texttt{VP}_{4} \hspace{1cm} & \texttt{S}_{i} \ra \texttt{NP} \ 
  \texttt{VP}_{i}
\end{array}
\end{equation}

None of these options is an incremental way of capturing exceptions in
a grammar. The introduction of an exception forces the revision of
previously described lexical types and grammar rules.

This naive example exemplifies a problem that becomes significant when
an operation or rule carries on most of the information-combining
task. This is a standard situation in lexicalist approaches to
grammar.  For instance, \cite{Dowty-88} proposes a distinction between
a basic categorial grammar, which only includes {\em forward} and {\em
  backward application}, and an extended grammar that also includes
{\em type raising} and {\em functional composition}. The basic grammar
is used for most of the analysis, whereas the extended one should only
be used in specific situations such as non-constituent coordination.
It is suggested that this rules should be licensed only when
functional application fails to parse a string. However, no
formalization of the different status of such rules is offered.

We propose a formalism to express and handle default and exceptional
rules within a phrase-structure approach that permits this kind of
representations.  Our formalism describes rules as object-oriented
functions that map daughters information into mother features. As in
object-oriented programming, precedence between rules is not given
statically in the linguistic signature, but it is dynamically
established upon the types of the input structures of the
compositional process. When combining two constituents, a
metagrammatical mechanism establishes a partial order for the
candidate rules and selects the most appropriate rule (or set of
rules) to be applied.

Before defining such framework, let us introduce the proposal with a
generic rule representation for the naive example above. The generic
rule having the desired behaviour for rule $\texttt{S} \ra \texttt{NP}
\ \texttt{VP}$ and rule $\texttt{S}_{i} \ra \texttt{NP} \ 
\texttt{VP}_{i}$ has this aspect:

\begin{equation}
  \label{SYN2}
    \grule{SYN} = \left\{ 
      \begin{array}{l} 
        \prule{SYN}{NP}{VP} (\type{x},\type{y}) = \type{S} \\ 
        \pprule{SYN}{\type{NP}}{\type{VP}_{i}} (\type{x},\type{y}) =
        \type{S}_{i}  
      \end{array}
      \right.
\end{equation}

Equation~\ref{SYN2} describes a generic rule, $\grule{SYN}$, that
returns a type for the composition of two linguistic objects.
$\prule{SYN}{NP}{VP} (\type{x},\type{y}) = \type{S} $ is a partial
rule applicable when the arguments $\type{x},\type{y}$ belong to the
types $\type{NP}, \type{VP}$, in that order.
$\pprule{SYN}{\type{NP}}{\type{VP}_{i}} (\type{x},\type{y}) =
\type{S}_{i} $ is applicable when the arguments belong to the types
$\type{NP},\type{VP}_{i}$.

A bottom-up parsing process may call $\grule{SYN}$ when it tries to
combine two objects with types $\type{x},\type{y}$. The applicable
partial rules will be $\{ \prule{SYN}{a}{b} \ | \ \type{a} \leq
\type{x}, \type{b} \leq \type{y} \}$. The natural way to establish
precedence between rules is attending to the specificity of the type
constraints of their arguments: $\prule{SYN}{a}{b}$ will be more
specific than $\prule{SYN}{c}{d}$ if $\type{a} \leq \type{c}, \type{b}
\leq \type{d}$.

Some examples of the application of $\grule{SYN}$ would be:
\begin{equation}
  \label{ejemplosSYN}
  \begin{array}[c]{ll}
    \grule{SYN}(\type{NP},\type{VP}) = \type{S} \hspace{1cm}& 
    \grule{SYN}(\type{NP},\type{VP}_{2}) = \type{S} \\ 
    \grule{SYN}(\type{VP}_{2},\type{NP}) = \nil & 
    \grule{SYN}(\type{NP},\type{VP}_{i}) =  \type{S}_{i}  
  \end{array}
\end{equation}

In each case, the following partial rules have been applied:
\begin{equation}
  \label{ejemplosSYN-2}
  \begin{array}[c]{ll} 
    \grule{SYN}(\type{NP},\type{VP}) \Longrightarrow
    \pprule{SYN}{\type{NP}}{\type{VP}} & 
    \grule{SYN}(\type{NP},\type{VP}_{2}) \Longleftrightarrow
    \pprule{SYN}{\type{NP}}{\type{VP}} \\ 
    \grule{SYN}(\type{VP}_{2},\type{NP}) \Longleftrightarrow 
    \nil \mbox{ (no applicable rule) }  & 
    \grule{SYN}(\type{NP},\type{VP}_{i}) \Longleftrightarrow
    \pprule{SYN}{\type{NP}}{\type{VP}}  
  \end{array}
\end{equation}
 
In that way, we have the ability to express default rules, and rules
with different degrees of specificity in an incremental way, without a
need to give them different formal status. Also, only minor changes
have to be made to a CFG parsing algorithm in order to work with
generic rules.

\mysubsection{Definition of a Generic Rule}
\label{definition}

Consider a partially ordered set ({\em poset}) $\poset{T}$ and a
domain of linguistic objects $\set{O}$ typed with $\poset{T}$ (i.e.,
there is a function \fun{type}:$\set{O} \rightarrow \set{T}$).
Consider also an informational domain $\Phi$ associated to $\set{O}$
by means of a function $\phi$:$\set{O} \rightarrow \Phi$ (perhaps
multivaluated) that associates at least one element in $\Phi$ to every
object of $\set{O}$.

A {\em generic rule} $\grule{G}$ over the tuple
$\tuple{\poset{T},\set{O},\Phi}$ is another tuple
  $\tuple{\cposet{T},\set{F},\dbfun{G}}$, where: 

\begin{itemize}
\item
$\set{F}$ is a set of functions,  that will be called  {\em partial
  rules},  of the form:
\begin{center}
$\prulef{f}{t} :\Phi \times \Phi \rightarrow \Phi$
\end{center}
where $\type{t}_{1}, \type{t}_{2} \in \set{T}$, and the following 
condition holds on the set of partial rules: \\
$\forall \prulef{f}{t},\prulef{f}{q}\in \set{F},
\type{t}_{1}=\type{q}_{1}\wedge 
\type{t}_{2}=\type{q}_{2} \Longleftrightarrow 
\prulef{f}{t}=\prulef{f}{q}$.  

\item
$\cposet{T}$ is a partially ordered poset defined over 
$\set{F}$ as follows:
\begin{center}
$\cset{T}\equiv\{ \vv{t} | \prulef{f}{t} \in \set{F} \}$ \\ 
$\forall \vv{x}, \vv{y} \in \cset{T}$ \ \  
$\vv{x}  \corder \vv{y}    
\Longleftrightarrow 
\type{x}_{1}  \leq \type{y}_{1}  \ and \   \type{x}_{2}  \leq
\type{y}_{2}$ \\ 
\end{center}  
 
We call {\em cartesian poset} $\cposet{T}$ over a poset $\poset{T}$ to 
any subset of $\cproduct{T}$ equipped with the partial ordering 
relation $\corder$. 
 
We call {\em cartesian type} of a partial rule $\prulef{f}{t}$ to the 
type specification $\vv{t}$. 
 
\item  
 
\begin{center} 
  $\forall \type{x}_{1},\type{x}_{2} \in \set{T}, 
  \dbfun{G}(\type{x}_{1},\type{x}_{2})\equiv \prulef{f}{t}$ such that 
  $\vv{t}$ is the lowest upper bound of $\vv{x}$ in $\cset{T}\cup 
  \{\vv{x}\}$. 
\end{center} 
 
$\dbfun{G}$ is called the {\em dynamic binding function} for 
$\grule{G}$. It interfaces parsing processes with the grammar, 
selecting the most specific partial rule in  $\set{F}$ given two 
arguments of types  $\type{x}_{1},\type{x}_{2}$. 
 
\end{itemize} 
 
Note that $\cset{T}$ is a subset of the cartesian product 
$\cproduct{T}$, and that the partial ordering relation $\corder$ is 
deduced from the relation $\leq$ that holds between the elements of 
$\set{T}$. By means of the definition of {\em cartesian poset}, we
have 
managed to axiomatize precedence issues between partial rules, turning 
the dynamic binding process into a simple order checking over the 
elements of a poset.  
 
The function $\dbfun{G}$ provides the interface of the generic rule, 
as a system of object-oriented functions, with the parsing process 
that calls it. In particular, the dynamic binding function provides 
the default interpretation for partial rules. Given a pair of 
arguments of types $\type{x}_{1},\type{x}_{2}$, a dynamic extension  
$\cset{T}\cup \{\vv{x}\}$ of the original cartesian poset $\cposet{T}$ 
that includes the cartesian type $\vv{x}$ is considered. In virtue of 
the definition of the partial ordering relation $\corder$, the new 
type $\vv{x}$ takes its place in the hierarchy. Its supertypes denote 
all the applicable rules to compose the pair of arguments, and the 
partial ordering between this supertypes denotes precedence between 
them. The action of the generic rule $\grule{G}$ over a couple of 
linguistic objects with types $\type{x}_{1},\type{x}_{2}$ and 
associated expressions $\phi_{1},\phi_{2}$ is given by 
$\dbfun{G}(\type{x}_{1},\type{x}_{2})(\phi_{1},\phi_{2})$.

In our definition, precedence is used to keep the most specific rule 
and override the rest. However, more sophisticated binding processes 
could be considered. For instance, if the partial rules were 
unification constraints, some default version of  unification 
could be performed on the 
applicable rules, taking their relative precedence into account. 
 
We have adopted the restriction of binary branching (binary rules) to 
take full advantage of the concept of dynamic binding with the minimum 
effort. However, this is not a serious limitation: most linguistically 
significant rules are binary, and those which are not can be easily 
converted in binary rules. 
 
\mysubsection{Well-formedness} 
\label{well-formedness} 
 
Given the definition of the preceding section, there can be situations 
where precedence conflicts cannot be solved. When 
the cartesian type that represents the arguments of a call to the 
generic rule has more than one direct supertype, there is not a single 
``most specific rule'' (see the example in Figure~\ref{her+earrings}).

\begin{figure} 
\hspace{3cm} 
$\poset{T}$ \hspace{4cm} 
$\langle \corder ,{\cal 
  T}^{*}\cup$\{\ctype{her}{earrings}\}$\rangle$ 
 
\hspace{-0.4cm} 
\begin{tabular}{ccc} 
                               & \node{a1}{\type{sign}}\\[2ex] 
\node{b1}{\type{pronoun}}   &                      &
\node{c1}{\type{noun}}\\[2ex] 
\node{d1}{\type{possessive}}  &                      &
\node{e1}{\type{count-noun}} 
\end{tabular} 
\nodeconnect{a1}{b1} 
\nodeconnect{a1}{c1} 
\nodeconnect{b1}{d1} 
\nodeconnect{c1}{e1} 
\hspace{-0.2cm} 
\begin{tabular}{ccc} 
\node{a}{\ctype{sign}{sign}}\\[2ex] 
\node{b}{\ctype{pronoun}{sign}} \\[2ex] 
\node{c}{\ctype{possessive}{noun}}  & &
\hspace{-1cm}\node{d}{\ctype{pronoun}{count-noun}}\\[2ex] 
                                                     &
                                                     
   \node{e}{\ctype{her}{earrings}} 
\end{tabular}
\nodeconnect{a}{b}
\nodeconnect{b}{c}
\nodeconnect{b}{d}
{\makedash{4pt}
\nodeconnect{c}{e}
\nodeconnect{d}{e}
\nodeoval{e}}
\ccaption{When the generic rule associated to $\cset{T}$ is invoked to 
  compose $\type{her}$ and $\type{earrings}$, a cartesian type 
  $\ctype{her}{earrings}$ is dynamically considered, but it has more  
  than one direct supertype; therefore, there is not a single most 
  specific rule to be applied on that constituents.} 
\label{her+earrings} 
\end{figure} 
 
If this indeterminacy is possible for a 
given generic rule, we say that the Cartesian Poset $\cposet{T}$ is 
not well-formed:  
 
{\em A Cartesian Poset $\cposet{T}$ 
is {\bf well-formed} iff \\ 
$\forall ~~ \vv{x},\vv{y} 
\in \cposet{T}$ not ordered, such that 
$\type{m}_{1}=\type{x}_{1} \wedge \type{y}_{1} ~  \in \poset{T}$ and  
$\type{m}_{2}=\type{x}_{2} \wedge \type{y}_{2} ~~  \in \poset{T}$, it
holds that 
$\vv{m} \in \cposet{T}$ 
} 
\footnote{ 
This definition assumes that the original poset is bounded complete, 
i.e., each pair of types have at most one common subtype.} 
 
This kind of indeterminacy appears also when dealing with multiple
default 
inheritance \cite{Daelemans-92} or default unification over feature 
structures with structure-sharing \cite{Bouma-92,Carpenter-93}, and 
can only be solved adding extra  
ordering information or forbidding such situations. Our 
well-formedness condition adopts the first solution. The main reason 
is that the cartesian hierarchies are not defined by the user, but 
arranged by the system. Such conflicts are potentially dangerous, as 
the grammar writer is not necessarily aware of them. An advantage of 
the definition of well-formedness is that it can detect 
inconsistencies at compile-time and signal them for correction.

\mysubsection{Parsing with Generic Rules} 
\label{parsing} 
 
A generic rule may interact with a phrase-structure grammar to perform 
the composition of some informational field. In \cite{Gonzalo-95}, for 
instance,  we propose generic rules as a well-suited mechanism to 
perform categorial semantic interpretation as a modular process that 
interacts with a phrase-structure grammar.  
On the other hand, a  grammar could be made up exclusively of generic 
rules, adequately combined to perform parsing.  
We will present a simple account of each of these possibilities.  
 
We will adopt here the deductive 
parsing approach described in \cite{Shieber-94}; it provides 
a neat account of parsing systems, simplifying our presentation. 
We will start from a bottom-up shift reduce algorithm as presented in 
\cite{Shieber-94}. We will gradually adopt this algorithm to capture 
generic rule parsing. 
 
Let $\alpha$ be a string of terminals $w_{i}$. Let 
$[ \alpha \bullet {}, j ]$ stand for $\alpha w_{j+1}\cdots w_n 
\derivestar w_1\cdots w_n$. In \cite{Shieber-94} a shift-reduce 
bottom-up deductive parsing system is expressed as the following 
calculus:

\begin{center} 
\begin{tabular}{l} 
 
{\bf Axiom:} \  \ $[ {} \bullet {}, 0 ]$ \\  
 
{\bf Goal:} \ \ $[ S \bullet {}, n ]$  
\end{tabular} \hspace{1cm} 
\begin{tabular}{ll} 
{\bf Inference Rules:}  \\ 
{\bf \quad Shift}       & 
        \( \oneover{[ \alpha \bullet {}, j ]} 
                   {[ \alpha w_{j+1} \bullet {}, j+1 ]} \)      \\  
{\bf \quad Reduce}      & 
        \( \oneover{[ \alpha \gamma \bullet {}, j ]} 
                   {[ \alpha B \bullet {}, j ]} 
                        \quad \mbox{ $B \rightarrow \gamma$} \) 
\end{tabular} 
\end{center}

This parser can be augmented to do semantic interpretation. If each 
rule has an associated function that related the meanings of the 
elements in the rhs with the meaning of the lhs element, the basic 
tuples are $[ \alpha \bullet {}, j, \phi ]$. The reduce rule carries 
on semantic composition:\footnote{The other rules are modified 
  trivially.} 
 
\begin{center} 
\begin{tabular}{ll} 
 
{\bf \quad Reduce}      & 
        \( \oneover{[ \alpha \gamma \bullet {}, 
          j {}, \Phi < \phi_{1} \cdots \phi_{|\gamma|}> ]} 
                   {[ \alpha B \bullet {}, j {}, \phi < f(\phi_{1}
                     \cdots \phi_{|\gamma|})> ]} 
                        \quad \mbox{ $B \rightarrow \gamma$, f} \) 
 
\end{tabular} 
\end{center}

where f is the semantic composition function associated to the rule $B 
\rightarrow \gamma$.  
 
We can augment the shift-reduce parser in a similar way to represent 
the interaction between a generic rule and a phrase-structure grammar.
We have to take account of
two differences between generic rule parsing and the
syntactic-semantic parser above:

\begin{enumerate}
\item The action of the generic rule has to be specified through the 
  dynamic binding function. There is not a partial rule associated to 
  each context-free rule.  
\item The interpretation of the context-free rules has to be slightly 
  modified to take into account the hierarchization of 
  non-terminals. A hierarchy is equivalent, in context-free terms, to 
  a set of unary rules in which every inmediate-ordering relation is 
  expressed writing a type as a left-hand side of a rule, and its 
  subtype as the right-hand side. Being these rules ``compiled'' into 
  the hierarchy, a context-free rule as  $ T \rightarrow T_1 T_2 $, 
  has to be interpreted as $T \rightarrow X_1 X_2$ such that $X_1 \leq 
  T_1, X_2   \leq T_2$.  
\end{enumerate}

The basic element in the logic that represents the parsing algorithm 
is the same as in the preceding case: $[\alpha {} \bullet {}, j , 
\Phi]$. Now, $\Phi$ is an expression that belongs to the informational 
domain associated to a generic rule $\grule{G}$.  
 
The reduce rule that takes account of the interaction of a 
generic rule with a context-free grammar is 

\begin{equation}
\label{generic-parsing}
\begin{tabular}{ll}

{\bf \quad Reduce}      &  \\ 
       \(  \oneover{[ \alpha \type{x}_{1} \type{x}_{2}  \bullet {}, 
          j {}, \Phi {} < \phi_{1} {}  \phi_{2} > ]} 
                   {[ \alpha A \bullet {}, j {}, \Phi {} {} 
                    < \dbfun{G}(\type{x}_1,\type{x}_2)
                    (\phi_{1},\phi_{2})> ]} \)
       & \( \myoneover{A \rightarrow B {} C {},\type{x}_1 \leq B
         ,\type{x}_2 \leq C}   
                      {\dbfun{G}(\type{x}_1,\type{x}_2)
                        (\phi_{1},\phi_{2}) \neq \nil} \)  

\end{tabular}
\end{equation}

The interaction with the generic rule is expressed in the term
$\dbfun{G}(\type{x}_1,\type{x}_2)(\phi_{1},\phi_{2})$. The left part,
$\dbfun{G}(\type{x}_1,\type{x}_2)$, performs dynamic binding,
returning the effective rule that will be applied on
$\type{x}_1,\type{x}_2$.

The side conditions on this rule are: one, that exists an applicable
syntactic rule $ A \rightarrow B {} C {},\type{x}_1 \leq B ,\type{x}_2
\leq C $. And two, that the generic function, applied over
$\phi_{1},\phi_{2}$, returns a positive result.  Note that it could be
the case that there is an appropriate syntactic rule, but the
additional restrictions imposed by the generic rule do not hold
\footnote{ This would be the case of semantic disambiguation.}.

It is interesting to remark that the dynamic binding process depends
on the particular objects in the Reduce rule, not only on the
conditions of the context-free rule. That makes impossible to
pre-attach an ``effective partial rule'' to each context-free rule at
compilation time, such that the binding process could be performed
off-line. This fact is reflected in the term
$\dbfun{G}(\type{x}_1,\type{x}_2)$.  If it were
$\dbfun{G}(\type{B},\type{C})$, binding could be done at compile time.
Though this behaviour introduces and additional complexity factor in
generic rule parsing, it allows the specification of semantic and
syntactic processes as independent mechanisms that interact modularly.

A particular case of generic rule is that in which the combination
functions simply return a type, as in our first example. In such a
rule, the role of every partial rule is similar to that of a
context-free rule, and the generic rule works as a context-free
grammar in which rules has a default interpretation. For this kind of
generic rules, that combine syntactic information, it is senseless to
consider their interaction with a context-free grammar, as they are
parsable by themselves. Such parsing does not differ substantially
{}from context-free parsing. The main restriction is that, due to the
functional interpretation of partial rules, bottom-up algorithms are
best suited that top-down mechanisms, that would require the
definition of inverse functions.

The reduce rule that takes account of syntactic parsing with a
generic rule can be described as follows:

\begin{equation}
\label{syn-parsing}
\begin{tabular}{ll} 
 
{\bf \quad Reduce}      & \\  
        \( \oneover{[ \alpha \type{x}_1 \type{x}_2 \bullet {}, j ]}  
                   {[ \alpha  
                     \dbfun{SYN}(\type{x}_1,\type{x}_2)
                     (\type{x}_1,\type{x}_2) \bullet {}, j ]} \)  
        &
        \mbox{$\dbfun{SYN}(\type{x}_1,\type{x}_2)
          (\type{x}_1,\type{x}_2) \neq \nil$ }  
\end{tabular} 
\end{equation}

The expression
$\dbfun{SYN}(\type{x}_1,\type{x}_2)(\type{x}_1,\type{x}_2)$, with both 
arguments repeated, may seem confusing. Note, however, that each
couple of arguments plays 
very different roles. The first couple are the arguments of the dynamic
binding function, from which an effective partial rule is
obtained. Such partial rule is then applied to the second set of
arguments. They are related to the informational domain associated to
the generic rule, which is, in this case, the type information
associated to the linguistic objects. 

The side condition, in this case, is precisely the one that licenses
the syntactic combining operation. If
$\dbfun{SYN}(\type{x}_1,\type{x}_2)(\type{x}_1,\type{x}_2)$ does not
return a positive result, then we cannot build a phrase out of the
arguments $\type{x}_1,\type{x}_2$.

The computational cost of parsing processes associated to this kind
of generic rules carrying on syntactic information is obviously the
same as the cost of parsing a context-free grammar. The only
difference between both processes is the way to select the appropriate
rule when the Reduce step is called. In context-free parsing, this
step involves looking up the available rules and matching the objects
involved with the right-hand side of the rules. In the worst case,
this process 
introduces a factor G in the overall complexity of parsing, being G
the size of the grammar. For a generic rule, the reduce rule involves
introducing a new type in the hierarchy of rules. Again, this implies
a factor G in the overall complexity, being G the number of partial
rules associated to the generic rule. The dependence with the length
of the string is obviously the same, as the surface behaviour of a
context-free grammar and a generic rule is exactly the same for
bottom-up parsing.

Another interesting case is when we have a generic rule to
specify syntactic restrictions, and another one to perform semantic
interpretation. It is easy to specify an algorithm to parse such
grammars from the preceding cases. 
Now the elements of the logic have the form \( [ \alpha \bullet {}, j, 
\Phi ] \), and the parser includes the following reduce rule:

\begin{equation}
\label{syn-sem-parsing}
\begin{array}{l}
\begin{tabular}{ll}
{\bf \quad Reduce} & \\
\end{tabular}
\\
\begin{tabular}{c}
        \( \oneover{[ \alpha \type{x}_1 \type{x}_2 \bullet {}, j {},
          \Phi \phi_1 \phi_2 ]}
                   {[ \alpha 
                     \dbfun{SYN}(\type{x}_1,\type{x}_2)
                     (\type{x}_1,\type{x}_2) \bullet {}, j {}, 
                      \dbfun{SEM}(\type{x}_1,\type{x}_2)(\phi_1,\phi_2)]}  \)  
          \(
          \myoneover{\dbfun{SYN}(\type{x}_1,\type{x}_2)
            (\type{x}_1,\type{x}_2) \neq \nil}  
            {\dbfun{SEM}(\type{x}_1,\type{x}_2)(\phi_1,\phi_2) \neq \nil} \)  
\end{tabular} 
\end{array}
\end{equation}

Let us see a (linguistically weird ) example. Consider the grammar
made up of the following generic rules:

\begin{equation}
\label{SYN}
    \grule{SYN} = \left\{ 
      \begin{array}{l} 
        \prule{SYN}{NP}{VP} (\type{X},\type{Y}) = \type{S} \\ 
        \pprule{SYN}{\type{NP}}{\type{VP}_{i}} (\type{X},\type{Y}) =
        \type{S}_{i}  
      \end{array}
      \right.
\end{equation}

\begin{equation}
\label{SEM}
    \grule{SEM} = \left\{ 
      \begin{array}{l} 
        \prule{SEM}{NP}{VP} (\type{X},\type{Y}) =
        sem(\type{X})(sem(\type{Y})) \\
        \pprule{SEM}{\type{Proper-Noun}}{\type{VP}}
        (\type{X},\type{Y}) =
        sem(\type{Y})(sem(\type{X})) \wedge
        (\lambda P.P(\sem{Pete}))(sem(\type{Y})) 
      \end{array}
      \right.
\end{equation}

The syntactic rule has already been considered in section~\ref{ideas}.
We assume the same hierarchy of that example, augmented with a new
type $\type{proper-noun} \leq \type{NP}$.  The semantic generic rule
takes account of a very special semantic issue, namely that Pete has a 
weak personality and imitates his friends in everything they do.
If we consider the phrases

\begin{itemize}
\item  \type{Betty} : $\lambda P. P(\sem{Betty})$ ; \type{Betty}
  $\leq$ \type{proper-noun}  
\item  \type{got+angry} : $\lambda x. \sem{ANGRY}(x)$ ;
  \type{got+angry} $\leq \type{VP}_{i}$ 
\end{itemize}

the application of the deductive system above to the string
`{\em Betty got angry}' would be as follows:

\begin{enumerate}
\item \( [ \bullet {}, 0 {}, {} ] \)
\item \( [ \type{Betty} \bullet {} , 1 , \lambda P. P(\sem{Betty})
  ] \) ({\bf Shift})
\item \( [ \type{Betty} \quad \type{got+angry} \bullet {}, 2 , 
  \lambda P. P(\sem{Betty}), 
  \lambda x.  \sem{ANGRY}(x) ] \) ({\bf Shift})
\item \( [ \type{S}_i {} \bullet {}, 2 , \sem{ANGRY}(\sem{Betty})
  \wedge 
  \sem{ANGRY}(\sem{Pete})] \) ({\bf Reduce}) 
\end{enumerate}

In the only application of the reduce rule, the following terms were
used: 

\vspace{0.1cm}
{\small 
\noindent
  $\dbfun{SYN}(\type{Betty},\type{got+angry})(\type{Betty},
  \type{got+angry}) =  
  \pprule{SYN}{\type{NP}}{\type{VP}_{i}}
  (\type{Betty},\type{got+angry})= \type{S}_i $ 

  \noindent $\dbfun{SEM}(\type{Betty},\type{got+angry})(\lambda P.
  P(\sem{Betty},\lambda x.  \sem{ANGRY}(x)) =$

 $\prule{SEM}{Proper-Noun}{VP}(\lambda P.
P(\sem{Betty},\lambda x.  \sem{ANGRY}(x))$ =  
$\sem{ANGRY}(\sem{Betty}) \wedge \sem{ANGRY}(\sem{Pete})$
}
\vspace{.1cm}

This example illustrates the point, stated before, that it is not
possible to attach a semantic rule to each syntactic rule at compile
time. It is, essentially, a dynamic binding process. 

A phrase-structure grammar with a one-to-one correspondence between
syntactic and semantic rules would need 12 pairs of rules to get the
same behaviour as the two generic rules above. The incrementality and
modularity of the generic rule approach is evident in this case.

\mysection{Non-Constituent Coordination and Categorial Grammar}
\label{ncc}

\ 
The general scheme for coordination corresponds to the conjunction of
constituents belonging to the same type: `{\em Nothing is certain,
  except death and taxes}', `{\em Take the money and run}', `{\em the
  long and winding road}'. Within the categorial grammar framework,
such cases are solved using the basic function application rules
(corresponding to the non-associative Lambek calculus in a
sequent calculus presentation). 
Nevertheless, there are cases of the so-called {\em non-constituent
  coordination}, where the conjoined expressions are not constituents
in the classical sense: `{\em John met Jane yesterday and Chris
today}', `{\em John read a book about linguistics on Monday and a
journal about computers on Tuesday}' (Left-node raising) , `{\em
John made and Peter painted a wooden chair} (Right-node
raising). The most common solution is to postulate extra-grammatical
levels of representation and/or special purpose parsing algorithms.

There have been a number of proposals within the categorial framework,
however,  that
deal with such phenomena at a grammatical level, extending the number of
rules or the set of basic operators. The combinatory rules of type
raising and functional composition \cite{Steedman-85}, for instance,
introduce associativity in the structural resources of the grammar.
\cite{Dowty-88} uses these rules to assign a category to the
coordinated conjuncts. With the exception of 
\cite{Wittenburg-90}, that proposes a (less intuitive) normal form for
such rules that avoid spurious ambiguity, all the proposals
suffer from intractability of the parsing task. Significantly, none of
them includes, to our knowledge, a formal differentiation between
default and exceptional rules. 

We will consider here a particularly simple account for
non-constituent coordination
phenomena based on the {\em sequence product} operator introduced in
\cite{Solias-93}. Its reformulation as a generic rule will show the
advantages of expressing default and exceptional grammar rules. 

In the non-constituent coordination examples above, we may consider
`{\em Jane yesterday}', \phrase{a book about linguistics on Tuesday},
etc, as being tuples of expressions belonging to the tuple product.
\footnote{ For these simplest cases of Non-constituent coordination we
  may also use Lambek's associative product. However the introduction
  of the sequence product brings advantages in more intricate examples
  of coordination like the ones considered in \cite{Solias-93}.
  Therefore we will use the sequence product in consideration of
  further extensions of this work.}. In this case the conjunction
scheme type $(x \setminus x)/x$ would substitute a sequence product by
the variable $x$.

In order to introduce this operator we need to extend the basic
string algebra of types by adding a tuple operation. Thus the
algebra would be $(L^{*}, +, \langle .,. \rangle)$, being + the
concatenative and associative operation and $\langle .,. \rangle$
the operation of tuple formation.

The model-theoretic definition for the sequence product is as
follows:
\begin{equation}
\label{(8)}
D(\tuple{A , B} ) = \{ < x_{1}, x_{2} > : x_{1} \in
D(A), x_{2} \in D(B) \}
\end{equation}
The sequent rules corresponding to \ref{(8)} are:
\begin{equation}
\label{(9)}
\frac {\Gamma \Rightarrow A \quad \Delta \Rightarrow B}  
      {\Gamma, \Delta \Rightarrow  \tuple{A , B} } R \tuple{}
\qquad
\frac {\Gamma, A, B, \Delta \Rightarrow C}
      {\Gamma, \tuple{A, B}, \Delta \Rightarrow C} L \tuple{}
\end{equation}
 
Examples with more than two elements in the sequence product will need a
generalization of the tuple operation. This generalization is
straightforward 
using the standard definition of n-tuple: $< x_{1},...,x_{n} > = < <
x_{1},...,x_{n-1} > x_{n} >$.  
Now we are able to account for a sentence
like \phrase{John read a book about linguistics on Monday  
and a journal about computers on Tuesday} by using a 3-tuple $\langle
np, pp, (np \setminus s) 
\setminus (np \setminus s) \rangle$. Right-node raising examples
proceed in a similar way.

This approach avoids using type-raising (a rule that can be applied on 
any category at any time), but still suffers from intractability, as
it is available for every combination of types. Again, the problem
relies on the exceptional status that should be given to the rules
that deal with non-canonical phenomena.

\mysection{A
  Generic Rule to Parse Non-Constituent Coordination}
\label{application}

The framework of generic rules offers a natural way to express the
grammar to deal with non-constituent coordination as an arrangement
of default rules, where each combination of types is performed
according to the most specific rule available.

The essential rules we want to express are:

\begin{itemize}
\item {\em By default, two types are combined using functional
    application and, if functional application is not applicable, they
    cannot be combined.} 
\item {\em When we try to combine two or more verbal complements and
    there is 
    a conjunction inmediately preceding them, a sequence product can
    be formed 
    with them.}
\item  {\em When we try to combine a noun phrase and a verb followed
    by a 
    conjunction, a sequence product can be formed with them}. 
\end{itemize}

``Verbal complement'', for our present purposes, stands for a
$\type{np}$, a $\type{vm}$ or, in turn, a sequence product
\footnote{We use
  $\type{vm}$ as an abbreviation for $(\type{np}\bk \type{s})\bk
  (\type{np} \bk \type{s} )$.  In this application we are only taken
  under consideration the possibility of having types $\type{np}$ and
  $\type{vm}$ as verbal complements. This set should be extended if
  some other kind of complementation were considered.}.

Such rules would guarantee that a sequence product is formed only in
the relevant 
cases. We only need an additional rule to scan optimally the elements
that match the sequence product on the left-side of the coordination. 

The formalism of generic rules allows for a direct specification
of such set of - still very informal - rules. Once turned into a
generic rule, they can be parsed with any bottom-up context-free
recognition algorithm (reformulated as the shift-reduce parsing in
section~\ref{parsing}).  

The first step is to express the operations related to the grammar as
binary rules: 

\begin{equation}
\label{rules}
\begin{array}{ll}
\fa : X/Y  \quad Y \ra \quad X  &
\Ituple : conj \quad  X \quad  Y  \ra \quad  conj \quad \tuple{X,Y}  \\ 
\ba : Y \quad Y\bk X \quad \ra \quad X \hspace{1cm} &
\Dtuple : \tuple{X_1 \cdots X_n} \ra X_1\cdots X_n \\
\end{array}
\end{equation}
\centerline{
$\begin{array}{l}
\scan : X_n \quad \tuple{\tuple{ X_1 \cdots X_{n -1}} \, X_n} \bk
\tuple{X_1 \cdots X_m} \ra \tuple{X_1 \cdots X_{n-1}} \bk \tuple{X_1
  \cdots X_m}
\end{array}$}

\

$\fa$ and $\ba$ are the usual forward and backward application rules.
$\Ituple$ is the rule to introduce a sequence product. As stated, it
seems a 
context-sensitive rule: The formation of a sequence product is only
possible when 
a conjunction is present to the left of the elements that form a
sequence product~\footnote{
As \cite{Solias-92} pointed out, the conjunction licenses type
sequences. Therefore, sequence product can appear in coordination 
environments, whereas other categories do not.}.
However, this
context-sensitivity does not overcome context-free grammar parsing, as 
the licensing element is a lexical item, a terminal, and its presence 
can be checked in constant time.  $\Ituple$ rule implements the 
stepped application of rules $L/$ and $R \tuple{}$ in the cancellation 
of type $(\langle A,\ B \rangle
\setminus \langle A,\ B \rangle) / \langle A,\ B \rangle \quad A,\ B$.
$\scan$ matches elements to the left of the
conjunction with the items in the tuple built to the right. This
reformulation is intended to a) keep the binary rules arrangement to
allow 
easy formulation of the generic rule, and b) take into account the
asymmetry between the formation process for the right and the left
coordinated conjuncts \footnote{The stepped cancellation of the right
  and the left conjuncts has well-known linguistic motivation, first
  stated in \cite{Ross-67}.} . The necessity of bulding a tuple is given
by the right conjunct, and the left conjunct is built only to match
the right one.  $\scan$ is implementing the consecutive application of
$ L \setminus$ 
and $R \tuple{}$. 
$\Dtuple$ is the rule to eliminate the
tuple operator and it implements the $L \tuple{}$ rule.

The crucial point to write a generic rule based upon the rules above
is to determine the types of the arguments associated to each
operation. By default, any combination of categories
is driven by functional application rules: forward
application ($\fa$) and backward application ($\ba$). Therefore, the
first 
partial rule  would be licensed on arguments of the most
general type $\type{T}$. This rule will try to apply $\fa$ or $\ba$
, and will return $\nil$ if both fail, forbidding that
combination.

The second partial rule is the rule of tuple formation. The type
conditions over the arguments of such a rule arise in a natural way
{}from the informal specification made at the beginning of this section:
both tupled elements have to be verb complements (in the case of
left-node raising) or an np followed by a verb (in the case of
right-node raising). We need two partial rules to establish both type
specifications. 
We will consider a
type $\type{C}$ that has $\type{np}$ and $\type{vm}$
as direct subclasses. Another possibility
is that one of the tupled members is, in turn, a tuple. Therefore, 
the tuple has to
be introduced as a direct subclass of $\type{C}$ as well. That
dependency provides the possibility of building n-tuples.

Finally, we need a scan rule to match objects at the left of the
conjunction once the right tuple has been combined with
the conjunction (by means of simple forward application). Therefore,
the type constraint on right elements is type $\type{C}$ again. 
The second one has to be 
a coordination of tuples missing some elements to its right. We will
denote this type as $\specialconj$. This
rule has to act in coordination with $\Dtuple$, which will be applied
only after the last conjoined element in the right coordinated
conjunct has been cancelled. 

To establish the interaction between the rules of
\ref{rules} and the partial rules, we will use the following
operations of composition, disjunction and optionality:

\begin{itemize}
\item $\comp{r}{p}(x,y) \equiv r(p(x,y))$ 
\item $\disj{r}{p}(x,y) \equiv \left\{ \begin{array}{l}
                             r(x,y) \mbox{ if } 
                             r(x,y) \neq \nil, p(x,y) = \nil \\
                             p(x,y) \mbox{ if } 
                             p(x,y) \neq \nil,  r(x,y) = \nil \\
                             \nil \mbox{ if } 
                             p(x,y) = \nil , r(x,y) = \nil 
                                       \end{array}
                                       \right.$
\item $\opt{r}(x) \equiv \left\{ \begin{array}{l}
                                  r(x) \mbox{ if } r(x) \neq \nil \\
                                  x \mbox{ if } r(x) = \nil \\
                                 \end{array}
                                 \right.$
\end{itemize}

That is the fragment of hierarchy that we need for our present
purposes:

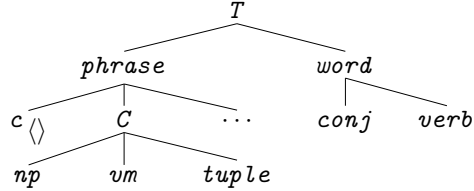
\begin{center}
\begin{tabular}{ccccc}
          &                  & \node{a}{\type{T}}  & \\[2ex]
          & \node{b}{\type{phrase}} &              &
          \node{c}{\type{word}}
\\[2ex]
\node{d}{\specialconj} & \node{e}{\type{C}}      & \node{f}{$\cdots$}
& \node{g}{\type{conj}} & \node{k}{\type{verb}}\\[2ex] 
\node{h}{\type{np}}     & \node{i}{\type{vm}} & \node{j}{\type{tuple}} &
\nodeconnect{a}{b}
\nodeconnect{a}{c}
\nodeconnect{b}{d}
\nodeconnect{b}{e}
\nodeconnect{b}{f}
\nodeconnect{c}{g}
\nodeconnect{e}{h}
\nodeconnect{e}{i}
\nodeconnect{e}{j}
\nodeconnect{c}{k}
\end{tabular}
\end{center}

The type $\type{C}$ represents verb complements, as introduced above. 
The type $\type{\specialconj}$ represents the combination of a
conjunction 
with a tuple as its right coordinated conjunct. It is needed to
specify the scan rule over the appropriate kind of objects. The type
$\type{tuple}$ is included as a verb complement, to allow formation
and coordination of n-tuples.

Given that hierarchy, we propose the following generic rule to parse
sentences including non-constituent coordination:

\begin{equation}
\label{NCC-SYN}
    \grule{SYN} = \left\{ 
      \begin{array}{l} 
        \prule{SYN}{T}{T} (\type{X},\type{Y}) =  (\disj{\fa}{\ba})
        (\type{X},\type{Y}) \\
        \prule{SYN}{C}{C} (\type{X},\type{Y}) = \Ituple
        (\type{X},\type{Y}) \\
        \prule{SYN}{np}{v} (\type{X},\type{Y}) = 
                  \Ituple (\type{X},\type{Y}) \\
        \pprule{SYN}{\type{C}}{\specialconj} (\type{X},\type{Y}) = 
        (\comp{\Dtuple}{\scan}) (\type{X},\type{Y})
      \end{array}
      \right.
\end{equation}

Note, again, that no precedence has to be defined by the grammar
writer to control the interaction of the rules. An easy, natural
analysis of the suitable arguments for each rule has implicitly
defined a partial ordering between them. 

The Hasse diagram of the cartesian poset associated to that generic
rule is:

  
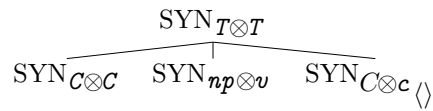
\begin{center}
\begin{tabular}{ccc}
                           & \node{a}{$\prule{SYN}{T}{T}$} & 
\\[2ex]
\node{b}{$\prule{SYN}{C}{C}$} & \node{d}{$\prule{SYN}{np}{v}$} &
\node{c}{$\pprule{SYN}{C}{\specialconj}$} 

\end{tabular}
\nodeconnect{a}{b}
\nodeconnect{a}{c}
\nodeconnect{a}{d}
  \end{center}

This rule can be parsed with the algorithm in~\ref{syn-parsing}. The
only novelty is that some word indexing has to be kept so that the
presence of a conjunction inmediately to the right of the arguments
can be checked in order to apply $\prule{SYN}{C}{C}$ and
$\prule{SYN}{np}{v}$. 

The Figure~\ref{fig:analysis} shows how dynamic binding works to get
an analysis of \phrase{John met Jane yesterday and Chris today}.
We have annotated each step of the analysis with the partial rule
effectively applied on the constituents being combined. That partial
rule is signalled by the dynamic binding function as the most specific 
to combine the types involved in the combination process.

\begin{figure}[tbp]
  \begin{center}
    \scriptsize{
\begin{tabular}{ccccccc}
John & met & Jane & yesterday & and & Chris & today \\
\node{a}{np} & \node{b}{(np\bk s)/np} & \node{c}{np} &
\node{d}{vp\bk vp} & \node{e}{(x\bk x)/x} &
\node{f}{np} & \node{g}{vp\bk vp}\\[2ex]
  &  &  &  &  &  & $\prule{SYN}{C}{C}$ \\[2ex]
  &  &  &  &  & \node{h}{$<$np, vp\bk vp$>$} & \\[2ex]
  &  &  &  &  &  & $\prule{SYN}{T}{T}$ \\[2ex]
  &  &  &  & \node{i}{$<$np, vp\bk vp$>$\bk $<$np, vp\bk vp$>$} 
& & \\[2ex]
  &  &  &  &  &  & $\pprule{SYN}{C}{\specialconj}$ \\[2ex]
  &  &  & \node{j}{$<$np$>$\bk $<$np, vp\bk vp$>$} & & & \\[2ex]
  &  & \node{empty}{} &  &  &  & $\pprule{SYN}{C}{\specialconj}$ 
\\[2ex]
  &  & \node{k}{np} &\node{k2}{ vp\bk vp} \\[2ex]
  &  &  &  &  &  & $\prule{SYN}{T}{T}$ \\[2ex]
  & \node{l}{(np\bk s)} \\[2ex]
  &  &  &  &  &  & $\prule{SYN}{T}{T}$ \\[2ex]
  & \node{m}{(np\bk s)}\\[2ex]
  &  &  &  &  &  & $\prule{SYN}{T}{T}$ \\[2ex]
\node{n}{s} & \\[2ex]
\end{tabular}
}
\nodeconnect{f}{h}
\nodeconnect{g}{h}
\nodeconnect{e}{i}
\nodeconnect{h}{i}
\nodeconnect{d}{j}
\nodeconnect{i}{j}
\nodeconnect{c}{empty}
\nodeconnect{j}{empty}
\nodeconnect{empty}{k}
\nodeconnect{empty}{k2}
\nodeconnect{k}{l}
\nodeconnect{k2}{m}
\nodeconnect{l}{m}
\nodeconnect{a}{n}
\nodeconnect{b}{l}
\nodeconnect{m}{n}
    \ccaption{Analysis of "John met Jane yesterday and Chris today" 
      according to $\grule{SYN}$.}
    \label{fig:analysis}
  \end{center}
\end{figure}

A context-free parser with the modifications shown in
section~\ref{parsing} for the {\bf reduce} step can produce this
analysis at a context-free cost in time, both in the length of the
string and the size of the grammar. Compared with the calculus
presented in section~\ref{ncc}, the generic rule does not suffer from
intractability and can be parsed with well-known, general parsing
techniques. 

A parser with heuristics or daemons to control coordination processes
could achieve a similar efficiency.  The clear advantage of a generic
rule is that the knowledge that reduces the search space is
declaratively introduced at the grammar level, and controlled at an
intermediate level between the grammar and the parser (by means of
dynamic binding).  This enhances linguistic motivation, modularity and
incrementality (both to extend the grammar and to control the parsing
processes).

\mysection{Conclusions}

The possibility of formally stating grammar rules with a default
interpretation has some advantages from a 
parsing perspective and from the point of view of 
knowledge representation.
On one side, it provides a declarative and modular way to reduce the
search space of parsing processes without altering parsing algorithms
with heuristic recipes.
On the other side, it
provides a linguistically motivated account of exceptional behaviour
that 
is particularly appealing for lexicalized grammar formalisms
where the lexicon is the repository of most of the linguistic
information, and there are only a few, very general rules that govern
linguistic phenomena. 

Our account of non-constituent coordination illustrates the advantages 
of such default arrangements for grammar rules. 
We have presented a categorial account of 
non-constituent coordination in which incombinable constituents on 
both sides of conjunction are 
treated as tuples of elements by the introduction of a sequence
type in the conjunction type. Once we have formed a single
tuple constituent, we can combine it with the remaining elements.
This approach of non-constituent coordination within Categorial
Grammar needs some rules of introduction and elimination
of the sequence operator which are not commonly needed for other 
simpler linguistic phenomena, and that makes the parsing process
intractable in its original Lambek-style formulation. 
When reformulated as a generic rule, the type conditions on the
arguments for each rule are used by the dynamic binding process to
fire the most appropriate grammar rule at every parsing step. The
process turns to be context-free parsable, and exhibits a highly
restricted search space. 

\newbox\partialpage
\def\starttwocolumn{%
  {\output={\global\setbox\partialpage=\vbox{\unvbox255}}\newpage}%
  \twocolumn[\unvbox\partialpage\vspace{14pt}]%
}
\starttwocolumn

\end{document}